\pdfoutput=1

\documentclass[11pt]{article}

\usepackage{emnlp2021}

\usepackage{times}
\usepackage{latexsym}
\usepackage{amsmath}
\usepackage{tabularx}
\usepackage{multirow}

\usepackage[T1]{fontenc}
\usepackage{fontawesome5}
\usepackage[T5]{fontenc}

\usepackage{tikz}

\usepackage[utf8]{inputenc}

\usepackage{microtype}

\title{Data Processing Matters: SRPH-Konvergen AI's Machine Translation System for WMT'21}

\author{Lintang Sutawika\thanks{\ \ Equal contribution. Order determined via coinflip.} \\
  Konvergen AI \\ 
  Jakarta, Indonesia\\
  \texttt{lintang@konvergen.ai} \\\And
  Jan Christian Blaise Cruz\footnotemark[1] \\
  Samsung Research Philippines \\
  Manila, Philippines \\
  \texttt{jcb.cruz@samsung.com} \\}

\begin{document}
\maketitle

\begin{abstract}
In this paper, we describe the submission of the joint Samsung Research Philippines-Konvergen AI team for the WMT'21 Large Scale Multilingual Translation Task - Small Track 2. We submit a standard Seq2Seq Transformer model to the shared task without any training or architecture tricks, relying mainly on the strength of our data preprocessing techniques to boost performance. Our final submission model scored 22.92 average BLEU on the FLORES-101 devtest set, and scored 22.97 average BLEU on the contest's hidden test set, ranking us sixth overall. Despite using only a standard Transformer, our model ranked first in Indonesian $\rightarrow$ Javanese, showing that data preprocessing matters equally, if not more, than cutting edge model architectures and training techniques.
\end{abstract}

\section{Introduction}
This paper describes the machine translation system submitted by the joint team of Samsung Research Philippines and Konvergen AI for the WMT'21 Large Scale Multilingual Translation Task. Our team participated in \textbf{Small Track \#2}, where the task is to produce a multilingual machine translation system for five Southeast-Asian languages: Javanese, Indonesian, Malay, Tagalog, and Tamil\footnote{Tamil is considered an official language in Singapore, a Southeast Asian country}, plus English, in all 30 directions.

We will first describe the filtering heuristics that we used to preprocess the data, and then outline the steps we took to train and evaluate our models. Specific hyperparameters, preprocessing decisions, and other training parameters will be listed in their corresponding sections. Finally, we report our results on the FLORES-101 \cite{goyal2021flores} devtest set, as well as on the competition's hidden test set. 

\section{Parallel Text Preprocessing Heuristics}

The contest dataset comprises of various bitext sources, including: 
bible-uedin \cite{christodouloupoulos2015massively}, 
CCAligned \cite{elkishky_ccaligned_2020}, 
ELRC 2922\footnote{https://elrc-share.eu/}, 
MultiCCAligned \cite{elkishky_ccaligned_2020},
ParaCrawl \footnote{https://www.paracrawl.eu/}, 
TED2020 \cite{reimers-2020-multilingual-sentence-bert}, 
WikiMatrix \cite{schwenk2019wikimatrix},  
tico-19, 
Ubuntu, 
OpenSubtitles, 
QED, 
Tanzil, 
Tatoeba, 
GlobalVoices, 
GNOME, 
KDE4, and 
WikiMedia \cite{TIEDEMANN12.463}.

We preprocess the datasets before training in order to minimize spurious relations that originate from incorrect text pairs. Our preprocessing removes samples based on a few heuristics that we developed based on our observation on the datasets. Each bitext file is applied a different set of preprocessing based on observation. For example we filter by number content for datasets such as CCAAligned while TED2020 is not applied that same filter.

In this section, we will cover the decisions made during preprocessing. We observe a score increase of 1.91 BLEU on our submission model when the preprocessing is applied. We report the total number of lines filtered from the bitext for all language pairs on Table \ref{tab:num_lines}.

\begin{table*}
    \centering
    \begin{tabular}{c|c|c|c|c}
        \hline
        \textbf{ISO} & \textbf{Language Pair} & \textbf{Before Preprocessing} & \textbf{After Preprocessing} & \textbf{Reduction} \\
        \hline
        en-id & English - Indonesian & 54,075,891 & 27,186,074 & 49.73\% \\
        en-ms & English - Malaysian & 13,437,727 & 7,674,956 & 42.89\% \\ 
        en-tl & English - Tagalog & 13,612,403 & 5,302,768 & 61.04\% \\
        en-jv & English - Javanese & 3,044,920 & 388,766 & 87.23\% \\ 
        en-ta & English - Tamil & 2,115,925 & 1,420,827 & 32.85\% \\ 
        id-ms & Indonesian - Malaysian & 4,857,321 & 3,371,777 & 30.58\% \\ 
        id-tl & Indonesian - Tagalog & 2,743,305 & 1,823,140 & 33.54\% \\ 
        id-jv & Indonesian - Javanese & 780,119 & 432,734 & 44.53\% \\ 
        id-ta & Indonesian - Tamil & 500,898 & 393,336 & 21.47\% \\ 
        ms-tl & Malaysian - Tagalog & 1,358,486 & 985,493 & 27.46\% \\ 
        ms-jv & Malaysian - Javanese & 434,710 & 250,070 & 42.47\% \\ 
        ms-ta & Malaysian - Tamil & 372,623 & 351,416 & 5.69\% \\ 
        tl-jv & Tagalog - Javanese & 817,146 & 544,233 & 33.40\% \\
        tl-ta & Tagalog - Tamil & 563,337 & 482,618 & 14.33\% \\
        jv-ta & Javanese - Tamil & 65,997 & 48,806 & 26.05\% \\

        \hline
    \end{tabular}
    \caption{Number of parallel text lines per language pair before and after applying preprocessing}
    \label{tab:num_lines}
\end{table*}

\subsection{Filter by Duplicate}

Duplication is present throughout the dataset. Table \ref{tab:duplication} outlines samples of duplication based on three distinct types:
\begin{itemize}
    \item \textbf{Duplicates within the same language}  Within a subset file of a designated language, multiple lines have the same string while the its counterpart may feature different translations.
    \item \textbf{Partial duplication}  The whole string of text in one language is present in its counterpart translation.
    \item \textbf{Duplication among parallel text}  Both source and target text line feature exactly the same string. While this may be correct for named entities, most of these duplication are short and can be non-informative.
\end{itemize}

\begin{table*}[hbt!]
    \centering
    \begin{tabular}{cc}
        \hline
        \multicolumn{2}{c}{\textbf{Duplicates within the same file}} \\
        \hline
        \hline
        \textbf{GNOME.en-tl.en} & \textbf{GNOME.en-tl.tl} \\
        \hline
        Error reading from file: \%s & Error sa pagbasa ng talaksang '\%s': \%s \\
        Error seeking in file: \%s & Error sa pagbasa ng talaksang '\%s': \%s \\
        Error closing file: \%s & Error sa pagbasa ng talaksang '\%s': \%s \\
        \hline
        \multicolumn{2}{c}{\textbf{Partial duplication}} \\
        \hline
        \hline
        \textbf{WikiMatrix.en-jv.en} & \textbf{WikiMatrix.en-jv.jv} \\
        \hline
        CJ E\&M Corporation. & Drama iki diprodhuksi déning CJ E\&M Corporation. \\
        New Orleans, Louisiana. & Lair ing New Orleans, Louisiana. \\
        Edward Thomas Hardy. & Jeneng dawané ya iku Edward Thomas Hardy. \\
        \hline
        \multicolumn{2}{c}{\textbf{Duplication among parallel text}} \\
        \hline
        \hline
        \textbf{OpenSubtitles.en-ta.en} & \textbf{OpenSubtitles.en-ta.ta} \\
        \hline
        Those who are invited will find the way. & Those who are invited will find the way. \\
        Gazelle, whose face the full moon forms: & Gazelle, whose face the full moon forms: \\
        Time has warned us never to approach her. & Time has warned us never to approach her. \\
        \hline
    \end{tabular}
    \caption{Examples of duplication based on three types}
    \label{tab:duplication}
\end{table*}

\subsection{Filtering by Language and Letters}

In algorithmically-aligned datasets such as CCAligned, some training examples are not in the list of contest languages. We find full text lines that are in Azerbaijani, Turkish, Arabic, and Japanese. To identify these languages, we use langdetect\footnote{https://pypi.org/project/langdetect/}. This filter works for sentences that are fully foreign. It is also the case that foreign letters that may refer to named-entity can be found in the dataset. We consider this to be allowable so long as the the foreign character string is present in both source and target text line. To filter this, we use AlphabetDetector\footnote{https://pypi.org/project/alphabet-detector/} and check if detected foreign letters are present in both text line.

\subsection{Filter by Specific Keywords and Symbols}

There are a number of cases where the translations are generally correct but also feature extra keywords that have no relation to the parallel text. These keywords are generally in English and are consistently present in a number of bitext datasets such as KDE4, GNOME, and Ubuntu.

\begin{table*}[hbt!]
    \centering
    \begin{tabular}{cc}
        \hline
        \textbf{KDE4.en-id.en} & \textbf{KDE4.en-id.id} \\
        \hline
        Task Scheduler & Penjadwal Tugas\underline{Comment} \\
        Configure and schedule tasks & Atur dan jadwal tugas\underline{Name} \\
        \hline
    \end{tabular}
    \caption{Example of translations that also have an extra keyword. Underlined text are keywords that are misplaced in correct translations.}
    \label{tab:dataset_list}
\end{table*}

Bitexts such as OpenSubtitles feature secondary information that relates to a particular scene (for example "\textit{(loud music playing)}"). These secondary information may be in parentheses to denote an action being done or to signify a song being played. These secondary information are not always available for each language. We opt to remove all lines that have these specific symbols. 

\subsection{Filtering Number Content}

We apply a filter to remove incorrect text lines in the bitext by checking if both source and target text lines feature the same numeric values such as date and quantities. Table \ref{tab:filter_by_number} shows that filtering by number can remove text lines that do not relate to one another as numeric values tend to translate the same. Due to the limited time allotted for the shared task, we opt to remove entirely parallel sentences that do not have matching numbers. We filter this by using regular expressions.

\subsection{Filtering by Length}

Text lines with very long lengths are generally not informative, we find most of these text lines consists of a list of names that would normally be found in a bibliography. We set an arbitrary max length of 500 characters for both source and target sentences.

\begin{table*}[hbt!]
    \centering
    \begin{tabular}{ccc}
        \hline
        & \textbf{MultiCCAligned.id-tl.id} & \textbf{MultiCCAligned.id-tl.tl} \\
        \hline
        \multirow{2}{*}{Removed} & Di. 13:00 - 17:30 & Mo. 13:00 - 18:00 \\
        & Di 24 nov. 10h – 18h & Sa 23 nov. 10h – 18h \\
        \hline
        \multirow{2}{*}{Kept} & (Terakhir diperbarui saat: 24/03/2020) & (Huling nai-update Sa: 24/03/2020) \\
        & Harga / \$: 1,2835 & presyo / \$: 1.2835 \\
        \hline
    \end{tabular}
    \caption{Incorrect translations can be easily identified by checking whether numeric values in both strings match. In the first example, the sentence pair was removed due to differing date and time. In the second example, the sentence pair was kept as we do not check punctuation for numerical values.}
    \label{tab:filter_by_number}
\end{table*}



\section{Experiments}

\subsection{Model Architecture}
For our submission, we wish to measure how much performance can be boosted by heuristics-based data preprocessing alone. Given that we anticipate most, if not all, submissions to the shared task will be transformer-based models, we opt to use the standard ``vanilla'' Sequence-to-Sequence Transformer \cite{vaswani2017attention} model with little-to-no changes. This lets us more clearly compare the performance boost of our filtering heuristics against the boost provided by a number of architecture augmentations and training tricks that other submissions might have.

In addition to using a standard Transformer model, we only train the model directly on our filtered bitext and do not make use of Backtranslation \cite{sennrich2015improving} for data augmentation. We also start from-scratch with models initialized using Glorot Uniform \cite{glorot2010understanding}, opting not to use massively-pretrained translation models such as M2M-100 \cite{fan2021beyond} as our starting checkpoint.

Following \citet{vaswani2017attention}, we produce two models: a base model and a large model. For the sake of simplicity, for the rest of the paper, we will refer to our models trained with our filtered data as \textbf{$\text{Base}_{\text{Heuristics}}$} and \textbf{$\text{Large}_{\text{Heuristics}}$}.

The hyperparameters used for our models are presented in Table \ref{tab:model-hyperparameters}.

\begin{table}
    \centering
    \begin{tabular}{lcc}
        \hline
        \textbf{} & \textbf{$\text{Base}$} & \textbf{$\text{Large}$}\\
        \hline
        Vocab Size & 37,000 & 37,000 \\
        Encoder Layers & 6 & 6 \\
        Decoder Layers & 6 & 6 \\
        Attention Heads & 8 & 16 \\
        Embedding Dim. & 512 & 1024 \\
        Feedforward Dim. & 2048 & 4096 \\
        Dropout & 0.1 & 0.3 \\
        Attention Dropout & 0.1 & 0.3 \\
        Pos. Embeddings & Sinusoid & Sinusoid \\
        \hline
        Parameters & 63M & 214M \\
        \hline
    \end{tabular}
    \caption{Model hyperparameter choices for the base and large Transformer variants.}
    \label{tab:model-hyperparameters}
\end{table}

\subsection{Data Preformatting and Tokenization}
Our models employ one single shared vocabulary for all languages and directions. We train our tokenizer using the SentencePiece\footnote{https://github.com/google/sentencepiece} library, limiting our vocabulary to 37,000 BPE \cite{sennrich2015neural} tokens, and training with a character coverage of $0.995$.

Before training the tokenizer, we first preformat the dataset into the format to be used for training later on. We append the source and target language's ISO-639-1 code enclosed in square brackets at the beginning of each sentence. For example:
\begin{equation*}
    \text{\texttt{[en]} \texttt{[tl]} Today is a sunny day.}
\end{equation*}
is the preformatted version of "Today is a sunny day." when translating from English to Tagalog. 

This preformatting is only done for the source sentences in the training dataset, while the target sentences are untouched.

For the purpose of training the tokenizer, the six language tokens (\texttt{[en]}, \texttt{[id]}, \texttt{[jv]}, \texttt{[ms]}, \texttt{[ta]}, and \texttt{[tl]}) are treated as special tokens to ensure that they will not be segmented later on.

\subsection{Training Setup}
We then compile our filtered, preformatted bitext and train our base and large models. During training, we limit all source and target sentences to a maximum sequence length of 150 subword tokens. All sentences that are much longer are truncated. 

Our models are trained using the Adam \cite{kingma2014adam} optimizer. Following \citet{vaswani2017attention}, we also use the ``Noam`` learning rate scheduler, linearly increasing the learning rate from $0$ for the first 8000 steps, then decaying afterward. We also set Adam's $\beta_2 = 0.998$ and use a label smoothing factor of $0.1$.

For batching, we accumulate tokens until we reach a maximum size of approximately 32,000 tokens per batch, an increase over the 25,000 tokens used in \citet{vaswani2017attention}. We then train the base model and the large model for 100,000 steps and 300,000 steps, respectively. All our models are trained on 8 NVIDIA Tesla P100 GPUs in parallel using the OpenNMT-py \cite{klein-etal-2017-opennmt} toolkit.

\subsection{Translation}
To generate translations using the model, we use Beam Search with beam size $5$ and apply an average length penalty of $0.6$. During generation, we limit all outputs to a maximum sequence length of 100, preemptively terminating generation if it begins to exceed this maximum length. We do not use sampling during translation, nor increase the temperature parameter as this induces randomness \cite{lopez2020simplifying}.

We test our experimental models on the FLORES-101 devtest set. We report our BLEU scores using the SPM-BLEU variant of SacreBLEU\footnote{BLEU+case.mixed+numrefs.1+smooth.exp+tok.spm\newline+version.1.5.0} \cite{post-2018-call}.

\section{Results}
After training our models and producing sample translations from the FLORES-101 devtest set, we compare the results of our two models with a number of baselines:
\begin{itemize}
    \item Transformers with No Heuristics -- These models are essentially identical with our Transformer models in terms of architecture, hyperparameters, and training setups, except the bitext they are training on are the raw training corpus given in the competition (i.e. the filtering heuristics were not applied on them). We train these models as an ablation experiment to be able to identify how much of the final performance is attributable to the filtering heuristics.
    \item M2M-100 615M -- This is the baseline given for the WMT'21 Large-scale Multilingual Translation Task Small Track 2 competition. This M2M-100 \cite{fan2021beyond} model was trained on CCMatrix and CCaligned with no further finetuning on the contest dataset.
    \item DeltaLM+ZCode -- This is the best performing model for the Small Track 2. The model is a finetuned version of the DeltaLM \cite{ma2021deltalm} encoder-decoder pretrained model.
\end{itemize}

All analyses and results within this section are based on the \textit{public devtest set} and not the contest's hidden test set, unless specified. A summary of the BLEU scores for all models and baselines are available on Table \ref{tab:model-results}.

\begin{table*}
    \centering
    \begin{tabular}{l|cccccc}
         \textbf{} & \textbf{$\text{Base}_{\text{Heuristics}}$} & \textbf{$\text{Large}_{\text{Heuristics}}$} & \textbf{$\text{Base}$} & \textbf{$\text{Large}$} & \textbf{M2M100} & \textbf{DeltaLM}  \\
         \textbf{} & \textbf{} & \textbf{} & \textbf{} & \textbf{} & \textbf{Baseline} & \textbf{+ZCode}  \\
         \hline 
         en$\rightarrow$id & 35.94 & 39.29 & 35.12 & 36.51 & 36.34 & 50.90 \\
         id$\rightarrow$en & 31.20 & 33.40 & 29.22 & 30.93 & 33.33 & 47.35 \\
         en$\rightarrow$jv & 21.53 & 23.57 & 16.95 & 20.98 & 15.06 & 27.70 \\
         jv$\rightarrow$en & 22.09 & 24.61 & 18.85 & 21.26 & 21.38 & 39.44 \\
         en$\rightarrow$ms & 31.36 & 36.93 & 36.63 & 38.60 & 32.63 & 46.77 \\
         ms$\rightarrow$en & 31.92 & 33.16 & 30.31 & 32.97 & 33.63 & 47.86 \\
         en$\rightarrow$ta &  9.15 & 10.64 &  8.78 &  9.68 &  4.24 & 35.48 \\
         ta$\rightarrow$en & 17.00 & 19.55 & 15.83 & 18.47 &  7.52 & 35.29 \\
         en$\rightarrow$tl & 26.91 & 33.23 & 27.87 & 27.56 &  9.95 & 40.52 \\
         tl$\rightarrow$en & 31.22 & 33.65 & 26.51 & 29.61 & 26.59 & 48.55 \\
         id$\rightarrow$jv & 23.18 & 23.91 & 21.41 & 22.30 & 15.86 & 23.35 \\
         jv$\rightarrow$id & 25.45 & 27.10 & 24.15 & 25.15 & 23.21 & 34.64 \\
         id$\rightarrow$ms & 30.58 & 33.94 & 28.38 & 33.01 & 29.32 & 38.30 \\
         ms$\rightarrow$id & 30.94 & 33.68 & 31.29 & 32.54 & 31.44 & 40.36 \\
         id$\rightarrow$ta &  7.04 &  7.88 &  6.78 &  7.09 &  1.44 & 29.61 \\
         ta$\rightarrow$id & 13.74 & 16.46 & 13.35 & 14.87 &  4.99 & 28.56 \\
         id$\rightarrow$tl & 23.32 & 25.27 & 22.30 & 23.23 &  9.32 & 33.56 \\
         tl$\rightarrow$id & 25.31 & 27.76 & 23.40 & 25.03 & 20.76 & 38.70 \\
         jv$\rightarrow$ms & 23.36 & 25.08 & 19.92 & 23.63 & 19.57 & 33.14 \\
         ms$\rightarrow$jv & 21.08 & 21.29 & 12.33 & 20.97 & 14.22 & 23.91 \\
         jv$\rightarrow$ta &  4.70 &  4.97 &  3.85 &  4.62 &  3.52 & 24.19 \\
         ta$\rightarrow$jv &  9.25 & 11.13 &  7.54 &  9.22 &  2.51 & 18.35 \\
         jv$\rightarrow$tl & 17.43 & 19.61 & 15.79 & 17.31 & 11.96 & 28.50 \\
         tl$\rightarrow$jv & 16.96 & 18.82 & 14.56 & 17.00 & 12.31 & 23.17 \\
         ms$\rightarrow$ta &  7.01 &  7.87 &  6.65 &  7.23 &  2.38 & 28.83 \\
         ta$\rightarrow$ms & 15.09 & 16.64 & 14.54 & 16.44 &  4.70 & 26.83 \\
         ms$\rightarrow$tl & 23.30 & 24.97 & 22.17 & 23.01 & 11.04 & 32.81 \\
         tl$\rightarrow$ms & 25.86 & 27.10 & 23.19 & 25.85 & 18.16 & 36.15 \\
         ta$\rightarrow$tl & 15.26 & 18.43 & 14.98 & 16.05 &  3.15 & 26.64 \\
         tl$\rightarrow$ta &  6.27 &  7.65 &  5.89 & 6.60 &  3.10 & 28.80 \\
         \hline
         Average & 20.78 & 22.92 & 19.28 & 21.01 & 15.46 & 33.94 \\
         \hline
    \end{tabular}
    \caption{Summary of BLEU scores on the FLORES-101 devtest set. The first two columns show the performance of our Transformer models trained with the data filtering heuristics. The next two columns show the same Transformer models, but trained on an unpreprocessed version of the training dataset. We also show the scores of the M2M-100 615M baseline model, as well as the best performing model (DeltaLM+ZCode) for the Small Track 2. $\text{Large}_{\text{Heuristics}}$ (column 2) is our final submission model for the contest.}
    \label{tab:model-results}
\end{table*}

\subsection{Transformer + Heuristics vs. Baselines}
We report the results of our $\text{Base}_{\text{Heuristics}}$ and $\text{Large}_{\text{Heuristics}}$ models against the M2M-100 615M model baseline as well as the best performing model for the shared task.

$\text{Base}_{\text{Heuristics}}$ scored an average BLEU of 20.78 on all 30 directions. On the other hand, $\text{Large}_{\text{Heuristics}}$ scored 22.92 average BLEU on all 30 directions, which is 2.14 BLEU points higher than the base model. Both models outperformed the M2M-100 615M baseline, with the base model giving a 5.32 BLEU improvement, and the large model giving a 7.46 BLEU improvement.

It is worth noting that, while the $\text{Base}_{\text{Heuristics}}$ outperforms the baseline on average, it fails to outperform it on four specific translation directions: en$\leftrightarrow$id and en$\leftrightarrow$ms. Note that it is these two language pairs that have the most number of training sentences in the training corpus.

The language pairs that benefit significantly from training on the contest dataset are language pairs that are of less volume than en$\leftrightarrow$id and en$\leftrightarrow$ms. This is likely due to these pairs being less-sampled in M2M100's training dataset, and thus were not as learned by the model compared to pairs with a higher volume of training data.

The same observations can be found when comparing the performance of $\text{Large}_{\text{Heuristics}}$ against the baseline model. $\text{Large}_{\text{Heuristics}}$ only marginally outperformed the baseline in one direction (id$\rightarrow$en, +0.07 BLEU), and marginally underperformed against the baseline in one direction (ms$\rightarrow$en, -0.47 BLEU). This higher performance for M2M-100 is likely due to the training method used in the model in addition to the size of the training corpora used. While M2M-100 is advantageous in these translation directions, the difference is only marginal, most likely owing to $\text{Large}_{\text{Heuristics}}$'s size which gives it higher capacity.

Both our transformer models and the baseline model are significantly outperformed by the DeltaLM+ZCode model, which is the best performing model in the competition. The best model outperforms our best model ($\text{Large}_{\text{Heuristics}}$) by a significant 11.02 average BLEU, and the baseline model by 18.48 average BLEU.

While DeltaLM+ZCode outperforms our model in terms of average performance, it is worth noting that our model -- a standard Transformer without any augmentations and training tricks -- managed to outperform DeltaLM+ZCode in one translation direction: id$\rightarrow$jv. 

$\text{Large}_{\text{Heuristics}}$ scored 23.91 BLEU while DeltaLM+ZCode scored 23.35 BLEU. While the difference is marginal (+0.56 BLEU), our model still outperforms the best model in this direction, which we attribute to the quality of our data preprocessing and filtering heuristics.

\subsection{Heuristics vs. No Heuristics}
To quantify how much our filtering heuristics contributed to the final performance of our models, we trained two additional models: both identical to our base and large transformer variants, except the training corpus used was not processed using our filtering heuristics. For these ablation experiments, we use the same BPE tokenizer that is used for our main transformer models (trained on the filtered data). This is to ensure full model equivalency. To prevent confusion, we will refer to these ablation models simply as \textbf{$\text{Base}$} and \textbf{$\text{Large}$} to differentiate them from our contest models \textbf{$\text{Base}_{\text{Heuristics}}$} and \textbf{$\text{Large}_{\text{Heuristics}}$}.

On average, both sizes of models performed worse when trained without the filtering heuristics. $\text{Base}$ scored 19.28 average BLEU on the devtest set, 1.5 points lower than $\text{Base}_{\text{Heuristics}}$. On the other hand, $\text{Large}$ scored average 21.01 BLEU, which is 1.91 points lower than $\text{Large}_{\text{Heuristics}}$.

It is interesting, however, that $\text{Base}$ outperformed $\text{Base}_{\text{Heuristics}}$ in two translation directions: en$\rightarrow$ms and ms$\rightarrow$id. This may indicate that the filtering heuristics work better for a certain subset of languages. We look towards exploring how filtering methods such as ours affect multilingual translation datasets in terms of balance and informativeness in the future.

On the other hand, $\text{Large}$ performed worse than $\text{Large}_{\text{Heuristics}}$ in all 30 directions. This may be due to the increase in total trainable parameters, as larger models need more data with higher quality to be effectively trained. 

\subsection{The Case of Tamil}
We observe that our models, including the other models on the shared task leaderboard, struggled with Tamil. X$\leftrightarrow$ta translation is on average much worse in terms of BLEU score compared to the other translation directions that do not involve it.

We hypothesize that this is due to two things. 

First, Tamil is the most underrepresented language in the shared task dataset, with X$\leftrightarrow$ta having the least amount of parallel text for every language X in the training set. This causes the model, to a certain extent, to underfit on directions that translate to or from Tamil. 

Second, Tamil is the only language in the shared task dataset that does not use the latin alphabet. Combined with the fact that it is the most underrepresented language in the dataset, there is a possibility that the model may have treated Tamil as noise during training. The observation that X$\rightarrow$ta performs worse on average compared to its inverse direction ta$\rightarrow$X lends more credence to this hypothesis. The model is not trained well to represent sentences in Tamil, and thus, struggles when generating Tamil translations.

Part of our planned future work includes identifying methods to improve translation in multilingual datasets where the alphabets used may be more than one. This is to improve translation to non-latin alphabet languages in future methods.

\subsection{Hidden Test Set Performance}
We also report the performance of our models on the shared task's hidden test set. We once more compare our results against the baseline M2M-100 model as well as the best performing DeltaLM+ZCode model.

Our final submission for the shared task was our $\text{Large}_{\text{Heuristics}}$ model, which performed with an average BLEU of 22.97 on the shared task's hidden test set. This is a marginal difference from it's devtest set score (+0.05 average BLEU). 

$\text{Large}_{\text{Heuristics}}$, unsurprisingly, still outperformed $\text{Base}_{\text{Heuristics}}$ (20.73 average BLEU, +2.24 improvement) and the baseline M2M-100 model (14.02 average BLEU, +8.95 improvement) in the hidden test set. The shared task's best performing model, DeltaLM+ZCode, still outperforms all other models in the hidden test set, scoring 33.89 average BLEU, a 10.92 improvement over our best model.

On the hidden test set, $\text{Large}_{\text{Heuristics}}$ still ranked first in the id$\rightarrow$jv translation direction, scoring 24.05 BLEU. This outperforms DeltaLM+ZCode's 23.79 BLEU (+0.26) and M2M-100's 15.33 BLEU (+8.72).

A summary of our model's performance on the hidden test set, as well as the baseline and best performing model, can be found on Table \ref{tab:model-hidden-results}

\begin{table}
    \centering
    \begin{tabular}{lccc}
        \hline
        \textbf{} & \textbf{Public} & \textbf{Hidden} & \textbf{Rank}\\
        \textbf{} & \textbf{Test} & \textbf{Test} & \textbf{}\\
        \hline
        M2M-100 615M  & 15.46 & 14.02 & 8 \\
        DeltaLM+ZCode & 33.94 & 33.89 & 1 \\
        $\text{Base}_{\text{Heuristics}}$   & 20.78 & 20.73 & - \\
        $\text{Large}_{\text{Heuristics}}$  & 22.92 & 22.97 & 6 \\
        \hline
    \end{tabular}
    \caption{Average BLEU scores on the contest's hidden test set. The $\text{Base}_{\text{Heuristics}}$ model is unranked as it was not submitted as our final model.}
    \label{tab:model-hidden-results}
\end{table}

\section{Conclusion}
In this paper, we described the translation systems submitted by the joint Samsung Research Philippines-Konvergen AI team for the WMT'21 Large Scale Multilingual Translation Small Track 2 shared task. We outline the filtering heuristics that we took to preprocess our data. We then train two models with a bitext preprocessed using our filtering heuristics, with our best model reaching an average BLEU score of 22.92 on the devtest set, and outperforming the baseline model by 7.46 BLEU points. In addition, we rank sixth in the contest leaderboard overall, scoring 22.97 BLEU on the hidden test set. 

We also reached first place for the id$\rightarrow$jv translation direction, beating all other more complex models, despite only using a standard transformer without any special augmentations and training tricks. This provides empirical evidence that data quality and preprocessing decisions weigh just as much, if not even more, than cutting edge model architectures and training techniques do.

\bibliography{anthology,custom}
\bibliographystyle{acl_natbib}

\end{document}